\documentclass{article}
\usepackage{spconf}
\usepackage{amsmath,amssymb,amsfonts}
\usepackage{algorithmic}
\usepackage{graphicx}
\usepackage{textcomp}
\usepackage{xcolor}
\usepackage{booktabs}
\usepackage{url}
\usepackage{etoolbox}
\usepackage{microtype}
\usepackage[colorlinks=true,linkcolor=blue,citecolor=blue,urlcolor=blue]{hyperref}
\makeatletter
\patchcmd{\section}{-3.5ex \@plus -1ex \@minus -.2ex}{-1.8ex \@plus -.3ex \@minus -.2ex}{}{\PackageError{paper}{Section spacing patch failed}{Check article.cls}}
\patchcmd{\section}{2.3ex \@plus.2ex}{0.7ex \@plus.2ex}{}{\PackageError{paper}{Section spacing patch failed}{Check article.cls}}
\patchcmd{\subsection}{-3.25ex\@plus -1ex \@minus -.2ex}{-1.5ex\@plus -.3ex \@minus -.2ex}{}{\PackageError{paper}{Subsection spacing patch failed}{Check article.cls}}
\patchcmd{\subsection}{1.5ex \@plus .2ex}{0.65ex \@plus .2ex}{}{\PackageError{paper}{Subsection spacing patch failed}{Check article.cls}}
\patchcmd{\subsubsection}{-3.25ex\@plus -1ex \@minus -.2ex}{-1.5ex\@plus -.3ex \@minus -.2ex}{}{\PackageError{paper}{Subsubsection spacing patch failed}{Check article.cls}}
\patchcmd{\subsubsection}{1.5ex \@plus .2ex}{0.65ex \@plus .2ex}{}{\PackageError{paper}{Subsubsection spacing patch failed}{Check article.cls}}
\patchcmd{\@makecaption}{\vskip 10pt}{\vskip 2.5pt}{}{}
\patchcmd{\@maketitle}{\vskip 2em}{\vskip 0.5em}{}{\PackageError{paper}{Title top spacing patch failed}{Check spconf.sty}}
\patchcmd{\@maketitle}{\par\vskip 1.5em}{\par\vskip -0.6em}{}{\PackageError{paper}{Title spacing patch failed}{Check spconf.sty}}
\patchcmd{\@maketitle}{\vskip 1.5em}{\vskip 0.5em}{}{\PackageError{paper}{Title-author spacing patch failed}{Check spconf.sty}}
\newcommand{\compactsection}{\@startsection{section}{1}{\z@}%
  {-1.4ex \@plus -.3ex \@minus -.2ex}{0.5ex \@plus .2ex}%
  {\normalfont\normalsize\bfseries}}
\newcommand{\compactsubsection}{\@startsection{subsection}{2}{\z@}%
  {-1.5ex \@plus -.3ex \@minus -.2ex}{0.6ex \@plus .2ex}%
  {\normalfont\large\bfseries}}
\makeatother
\def\BibTeX{{\rm B\kern-.05em{\sc i\kern-.025em b}\kern-.08em
    T\kern-.1667em\lower.7ex\hbox{E}\kern-.125emX}}

\title{A FRONTEND-BACKEND ARCHITECTURE FOR TOOL CALLS IN FULL-DUPLEX SPEECH MODELS}

\name{\shortstack{Ke Hu, Slyne Deng, Chen Chen, Elena Rastorgueva, Edresson Casanova, \\
Punit Kumar, Dharmendra Choudhary, Nikhil Srihari, Ameya Sunil Mahabaleshwarkar, \\
Viet Anh Trinh, Slim Essid, Oluwatobi Olabiyi, Zhehuai Chen}}
\address{NVIDIA, USA}

\begin{document}
\ninept
\renewcommand{\baselinestretch}{0.87}\normalsize

\maketitle

\begin{abstract}
Full-duplex speech-to-speech (S2S) models provide natural, low-latency conversational interaction and would benefit from the ability to use external tools and complete voice-agent tasks. We propose a frontend-backend architecture in which a duplex speech-to-text frontend learns to emit a delegation token and forwards streaming ASR transcripts to a text-based backend LLM for tool calls. Tool-call results from the backend are injected back into the frontend through a lightweight \emph{prefill-and-repeat} mechanism and then synthesized using streaming TTS for the user. Our approach largely preserves regular duplex turn-taking, interruption handling, and low-latency interaction while requiring minimal modifications to the frontend model. In a single-turn tool-call evaluation, our system achieves 92--97\% tool-call recall, competitive tool-call prediction performance, and 81.2\% accuracy in rejecting irrelevant calls. When equipped with a larger backend (e.g., Qwen3-235B-A22B), our system achieves competitive results on Full-Duplex-Bench-V3 compared with open- and closed-source models and outperforms GPT-realtime-mini and Qwen3-Omni-30B-A3B-Instruct on EVA-A and EVA-X for EVA-Bench. These results demonstrate that backend delegation is an effective and modular approach for combining natural duplex speech interaction with strong agentic tool-call capabilities.

\end{abstract}


\section{Introduction}

 Full-duplex speech-to-speech (S2S) models are natural and desirable interfaces for conversational AI \cite{gptrealtime,gemini_live_api,xai_grok,moshi,personaplex,salmduplex,covoaudio, nemotron_voicechat_ea,nemotronlabs_voicechat_11b}. By operating directly in the speech modality, these models eliminate cascade latency, preserve paralinguistic cues, and produce natural turn-taking and barge-in behavior \cite{fdbench,fdbenchv2,easyturn,humdial}. While duplex speech models excel in natural conversation based on internal intelligence, text-based LLMs now serve as reliable tool-using agents: instruction-tuned models decide when and how to invoke external tools, query knowledge bases, and act on the world \cite{toolformer,gorilla,react,taubench,openai_gpt55,gemini_35_flash}.

Although there are increasing efforts to enable spoken tool calls in speech LLMs \cite{audio2tool,kimiaudio,qwen3omni,gptrealtime,geminilive3.1}, tool-call capabilities in voice agents remain substantially behind those of text-based agents. As shown in $\tau$-Voice \cite{tauvoice}, leading commercial duplex voice models complete only $31{-}51\%$ of grounded customer-service tasks under clean conditions, whereas text agents such as GPT-5 achieve $85\%$ on the corresponding text-mode tasks; the gap widens further under realistic noise and accented speech. This disparity suggests that natural spoken interaction and strong tool-call intelligence remain largely separate strengths of current systems. A natural question is therefore whether duplex speech models should directly internalize tool-call capabilities or instead delegate such capabilities to a backend text agent that already benefits from mature instruction following, tool calls, and long-horizon reasoning abilities.

Recent work investigates tool-call capabilities directly inside Moshi-style \cite{moshi} duplex speech models by placing tool calls in a separate channel \cite{duplexsla}. However, audio-native modeling imposes a fundamental capacity tradeoff: audio tokens consume parameters and context budget that text-only LLMs can devote to factual knowledge, instruction following, and tool-call capabilities \cite{kame}. In contrast, a delegation-style backend agent is more modular and consumes little modeling capacity from a frontend speech model.

In this direction, hybrid approaches that keep an S2S frontend for interaction while delegating tool calling and reasoning to a text backend appear in several concurrent designs: KAME \cite{kame} injects backend ``oracle'' tokens into a Moshi-style \cite{moshi} S2S frontend for knowledge integration; MoshiRAG \cite{moshirag} augments Moshi with retrieval; Thinking Machines \cite{thinkingmachine} proposes an interaction-background framework for user queries requiring deeper reasoning, tool calls, or long-horizon work. More recently, Qwen-audio-agent \cite{qwen_audio_agent} and GPT-Live \cite{gptlive} also adopt similar frameworks. However, it remains unclear how the delegation signal in \cite{thinkingmachine,qwen_audio_agent,gptlive} is designed or how the background model interacts with the duplex frontend.

In this work, we propose a frontend-backend architecture for executing tool calls and agentic tasks for a full-duplex speech model. Our S2S model consists of a duplex speech-to-text (STT) component (based on \cite{salmduplex,asru_duplex}) and a streaming TTS model \cite{eartts}. The duplex STT model takes encoded user speech, agent text, and streaming user ASR transcripts as inputs \cite{asru_duplex}. We use the duplex STT model as the frontend: it predicts a delegation token for voice queries involving tool calls and then sends the streaming ASR transcripts to the backend LLM, which handles the tool call in a LangGraph framework \cite{langgraph}. The backend's natural-language result is then sent back to the frontend agent-text channel via a \emph{prefill} mechanism, and the frontend is trained to repeat the result. The frontend remains silent during the tool calls and resumes its normal duplex behavior after the tool call is completed. Our design imposes minimal changes on the frontend duplex STT model by adding a tool-call control token to its prediction targets.
In a single-turn tool-call evaluation based on a speech version of BFCL \cite{servicenow_bfcl}, our experiments show that the frontend-backend architecture reliably predicts the tool-call token with 92\% to 97\% recall and rejects up to 81.2\% of irrelevant calls.
On more naturalistic user queries with pauses, hesitations, and self-corrections in Full-Duplex-Bench-V3 \cite{fdb3_hf}, our model with the Qwen3-235B-A22B backend achieves response quality similar to that of GPT-realtime-mini. 
On the more challenging EVA-Bench voice-agent task evaluation \cite{evabench}, which involves multi-turn conversations and tool calls, our model outperforms GPT-realtime-mini and Qwen3-Omni-30B-A3B-Instruct \cite{qwen3_30b_hf} on EVA-X and EVA-A and performs similarly to Gemini 3.1 Flash Lite \cite{gemini_31_flash_lite}.
Demos of our frontend-backend system can be found online.\footnote{\url{https://huggingface.co/spaces/frontend-backend-duplex/demo}}

\section{Related Work}

Cascaded frontend–backend systems, such as NVIDIA’s Nemotron Voice Agent \cite{nvidia_frontend_backend} and LiveKit’s EXA Deep Researcher \cite{livekit_exa_researcher}, combine ASR, LLM, and TTS into a frontend and use a separate backend for planning and tool execution. Recent systems increasingly explore the integration of tool use and backend agents into real-time voice interaction. Commercial systems \cite{gptlive, gptrealtime,gemini_live_spark} expose tool calls inside an end-to-end voice API, but their architectures are not disclosed. Recent work focuses on pairing a low-latency full-duplex speech frontend with a more capable text backend. KAME \cite{kame} runs a Moshi-style frontend and conditions its output on ``oracle'' tokens streamed from a backend LLM updated every 100--500\,ms; the frontend is explicitly trained to anchor its speech on these oracle tokens. MoshiRAG \cite{moshirag} augments Moshi with retrieval, injecting retrieved context into the inner-monologue stream. Thinking Machines adopts an interaction-background style system to delegate deep reasoning, tool calls, or longer-horizon jobs to the background model \cite{thinkingmachine}. Recently, Qwen-audio-agent \cite{qwen_audio_agent} has been proposed as a real-time voice model with backend agents that can execute various tasks while the frontend is in a conversation with the user. However, in \cite{thinkingmachine,qwen_audio_agent}, it is unclear how the delegation signal is designed or how backend results are integrated into the frontend duplex model.

Compared with the aforementioned approaches, we propose a straightforward approach that predicts a delegation token in the agent-text channel to assign tool calls to a backend. The backend executes multiple rounds of tool calls and then prefills the response text into the frontend as context for further generation. The proposed frontend-backend framework is modular and does not require significant architectural changes to the frontend.

\section{Architecture}

\subsection{Speech-to-text frontend}
\label{sec:frontend}

\begin{figure}[t]
\centering
\includegraphics[width=0.65\linewidth]{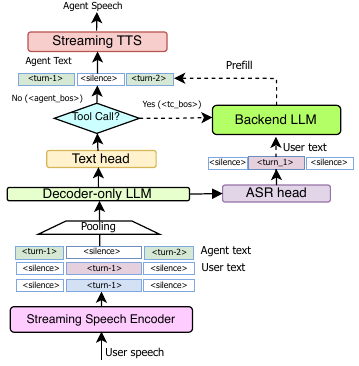}
\caption{The proposed frontend-backend system for a duplex speech-to-speech model with tool-call capability.}
\label{fig:fb}
\end{figure}
\vspace{-1ex}

As shown in Fig.~\ref{fig:fb}, our frontend model is a duplex speech-to-text (STT) model that consists of a streaming speech encoder and the LLM backbone (similar to \cite{hu2026enable}). The duplex STT model takes three input streams: user speech, user transcripts, and agent text. User speech is encoded by a 600M-parameter Parakeet streaming speech encoder \cite{parakeet} and fed to a backbone LLM (NVIDIA Nemotron-Nano-9B-v2-Base \cite{nvidia2025nemotron}). The streaming ASR head consists of a separate embedding layer and prediction head and shares the same LLM backbone as the agent-text head. A single decoding pass jointly produces both user and agent text.

To delegate tool calls to the backend, we train the frontend model to fire a tool-call token \texttt{<tc\_bos>} in the agent-text channel if the user's speech naturally requires a tool call (e.g., ``What is the weather in New York?''). We then train the model to generate a short filler (e.g., ``Let me pull that up.'') followed by the special token \texttt{<tc\_eos>}. Actual tool calls are performed by the backend by sending the user transcripts to the backend agent (see more details in Sect. \ref{sec:backend}), while in training we directly prefill the tool-call response from the training data into the agent-text channel and train the frontend to reproduce it exactly. The tool-call tokens, filler, and reproduced regions are used for loss computation in the same way as regular agent text, while prefill regions are masked and excluded from loss computation. During inference, we send the streaming ASR transcript to the backend for tool-call execution when \texttt{<tc\_eos>} fires. The filler allows time for the streaming ASR \cite{hu2026enable} to complete for full user transcripts. The backend returns the actual natural-language text for prefilling into the frontend (see Sect. \ref{sec:backend} for details). Our frontend is connected to VoiceChat-TTS \cite{eartts} to generate agent speech, which accepts explicit turn-start and interruption-control tokens and incrementally generates codec-based speech tokens.

\subsection{Backend agent}
\label{sec:backend}

We use LangGraph~\cite{langgraph}, a tool-augmented ReAct agent~\cite{react}, as our backend.
Our graph is a state machine over a running message list, comprising an agent node that employs an instruction-following LLM (e.g., \cite{qwen3_30b_hf}) and a tools node that executes the model's tool calls. The two nodes are linked by a conditional edge that routes to tool execution whenever the model emits calls and terminates the turn once the turn contains no tool calls. When the user's query requires a tool call, e.g., ``What is the weather in New York?'', the user ASR transcript is sent to the graph as a message, and the agent is invoked to call tools to respond. Multi-turn context is maintained automatically by a thread-keyed checkpointer that persists and restores the accumulating conversation state across turns, so each turn only takes the current user ASR transcript as input.
If the backend generates a syntactically invalid tool-call request or a regular text response, we return the text directly to the frontend as a prefill. This means the frontend may have incorrectly triggered a tool call, while the larger backend reliably falls back to natural-language explanations for irrelevant queries.

\section{Training and Inference}



For the frontend duplex STT model, our training includes a pretraining stage followed by supervised fine-tuning (SFT) \cite{hu2026enable}.
To enable tool-call token prediction in the frontend, we generate multi-turn conversation data with tool calls for SFT. First, we use various LLMs (e.g., Nemotron 3 Nano \cite{nemotron3nano}, Gemma-4-31B-IT \cite{gemma4_31b_it}, and Qwen3.5-397B-A17B \cite{qwen35_397b_a17b}) to generate user, agent, tool-call, and tool-response turns in text and use LLM judges to filter conversations with inconsistent turns or incorrect tool invocations. These textual conversations are then synthesized into audio conversations using a data pipeline involving filtering (e.g., removing math-heavy and code-heavy conversations not suitable for audio), TTS (VoiceChat-TTS \cite{eartts} with diverse voices), and ASR-based quality filtering (Parakeet-tdt-0.6b-v2 \cite{parakeetv2}) using WER/CER. The multi-turn tool-call data have variants targeting voice-agent decisions about when to call a tool, when to ask follow-up questions, and when not to call an unavailable or inappropriate tool, similar to When2Call \cite{when2call} but with multi-turn spoken user input. Our pipeline also simulates realistic multi-turn interactions with backchannels, pauses, and interruptions.
We have also constructed domain-based multi-turn tool-call data (e.g., simulated airline and retail databases) with entities distinct from those used in evaluation. We generate conversation and tool-calling trajectories by having two text-only LLMs interact with each other (both are Qwen3-235B-A22B \cite{qwen3_235b_hf}) to achieve diverse user goals. Samples that failed to complete the task are discarded, and we synthesize the user turns into speech using Chatterbox \cite{chatterboxtts2025}. In total, our training data include around 530k hours of pretraining data, 111k hours of SFT data, around 16k hours of ASR transcription data, and 8.5k hours of multi-turn conversation data with tool calls. When training the frontend, we randomly choose either the full tool definition as the system prompt or a partial prompt with only the function name and description for generalization. 

In training, if a user query leads to a tool call, we label the following agent turn as a tool-call turn, along with the final natural-language tool-call response. We first replace the regular \texttt{<agent\_bos>} token, which is usually placed 320\,ms after the end of the user turn (as in \cite{hu2026enable,salmduplex,asru_duplex}), with a \texttt{<tc\_bos>} token. To trigger tool calls more reliably, we add around 320\,ms of latency before \texttt{<tc\_bos>}, bringing the modeled filler response latency for tool-call turns to around 640\,ms. A filler lasting approximately 1\,s follows and ends with \texttt{<tc\_eos>}. The natural-language tool response is then enclosed by \texttt{<pf\_bos>} and \texttt{<pf\_eos>} and prefilled into the agent-text channel around 1\,s after \texttt{<tc\_eos>} to simulate the tool-call execution time, and the frontend model is trained to generate the exact prefilled text starting with \texttt{<agent\_bos>}. During training, we insert pad tokens into the ASR channel and silence into the speech regions used as inputs with prefilled agent text, and these regions are not used for loss computation. An example training sequence looks like:

\newcommand{\tok}[1]{\texttt{\textless{}#1\textgreater{}}}
\newcommand{\padrun}[1]{\tok{pad}\textsubscript{#1}}
\newcommand{\seg}[2]{[\,#1\,]\textsubscript{#2}}
\par\noindent\begin{minipage}{\linewidth}\fontsize{7pt}{8.5pt}\selectfont\raggedright
\textbf{User:} What is the weather in New York?\\[2pt]
\textbf{Agent:} \padrun{640\,ms} \seg{\tok{tc\_bos} Give me a moment. \padrun{}... \tok{tc\_eos}}{1\,s} \padrun{1\,s} \tok{pf\_bos} The weather in New York is 72 degrees \tok{pf\_eos} \tok{agent\_bos} The weather in New York is 72 degrees \padrun{}...\par\end{minipage}\par

The frontend is initialized from a checkpoint similar to the Nemotron VoiceChat model~\cite{nemotron_voicechat_ea} and fine-tuned for around 9k steps. We use AdamW with a learning rate of $3\times10^{-5}$, $\beta=(0.9, 0.98)$, no weight decay, and inverse-square-root annealing with 1k warmup steps. We use a minimum learning rate of $5\times10^{-6}$ and gradient clipping at 5.0.
During inference, once \texttt{<tc\_eos>} is detected, the frontend sends the user transcript, endpointed by \texttt{<user\_eos>}, to the backend to complete the tool-call request. The resulting natural-language response is then injected into the agent-text channel before further generation. During the function call, we suppress the agent output by inserting pad tokens into the agent-text channel. Our backend LLMs operate in non-thinking mode during all evaluations to reduce response latency.

\section{Results}

\subsection{Tool Calls}

\subsubsection{Single-Turn Tool Calls}

\begin{table}[!t]
\centering
{\footnotesize\caption{BFCL AST accuracy (\%) and irrelevance score. UV0.6-8B and UV0.6-32B denote Ultravox-v0.6 with Llama-3.1-8B and Qwen-3-32B backbones, respectively. Ours-7B uses Qwen2.5-7B, Ours-30B uses Qwen3-30B-A3B.}\label{tab:bfclv3}}
\footnotesize
\setlength{\tabcolsep}{1.5pt}
\setlength{\aboverulesep}{1.2pt}
\setlength{\belowrulesep}{1.2pt}
\renewcommand{\arraystretch}{1.0}
\begin{tabular}{@{}p{0.29\columnwidth}rrrrrr@{}}
\toprule
Model & Simple & Multiple & Parallel & Para. Multi. & Irrelev. & Avg \\
\midrule
GPT-realtime & \textbf{82.9} & \textbf{83.6} & \textbf{72.5} & \textbf{74.0} & \textbf{90.8} & \textbf{80.8} \\
\midrule
UV0.6-8B & 61.4 & 63.5 & 48.5 & 43.0 & 0.0 & 43.3 \\
UV0.6-32B & \textbf{78.9} & \textbf{81.1} & \textbf{75.2} & \textbf{65.4} & \textbf{82.5} & \textbf{76.6} \\
\midrule
Ours-7B-intASR & \textbf{80.2} & \textbf{79.6} & 69.5 & 52.7 & 76.7 & 71.7 \\
Ours-30B-intASR & 79.4 & 78.3 & 71.1 & 55.1 & \textbf{81.2} & 73.0 \\
Ours-30B-extASR & 80.1 & 78.7 & \textbf{72.0} & \textbf{61.1} & \textbf{81.2} & \textbf{74.6} \\
\bottomrule
\end{tabular}
\end{table}

We use the audio versions of the Berkeley Function-Calling Leaderboard (BFCL)
\cite{bfcl} single-turn datasets from ServiceNow-AI \cite{servicenow_bfcl} for evaluation.
The single-turn datasets include Simple, Multiple, Parallel,
 Parallel-Multiple, and Irrelevance subsets. We use the Abstract Syntax Tree (AST) score to evaluate
the structural correctness of the generated tool-call request. We follow the original AST
implementation \cite{bfcl_leaderboard}, which tolerates argument order, spacing, and formatting differences, as well as default parameters, in AST scoring.

As shown in Table \ref{tab:bfclv3}, we evaluate our system with backend LLMs of two different
sizes: Qwen2.5-7B~\cite{qwen25_7b_hf} and Qwen3-30B-A3B~\cite{qwen3_30b_hf}. We also compare two
methods for transcribing user speech: our internal ASR (intASR) transcripts and an external ASR (extASR) model \cite{parakeet}. 
The external ASR runs on the buffered user turn and adds a median latency of 0.30\,s per turn. As shown in Table \ref{tab:bfclv3}, we first find that the two backends with intASR perform similarly on Simple and Multiple, while the larger backend performs better on the Parallel and Parallel-Multiple subsets and rejects irrelevant speech more reliably. When we switch to external ASR transcripts for the Qwen3-30B-A3B backend, the average score improves from $73.0\%$ to $74.6\%$, with the biggest gain on Parallel-Multiple ($55.1\%\!\to\!61.1\%$) due to more accurate ASR transcripts. We will use the external-ASR setup in later experiments.

We also compare to GPT-realtime \cite{gptrealtime}, Ultravox-v0.6 with Llama-3.1-8B~\cite{ultravox_v06_llama31_8b} and Qwen-3-32B~\cite{ultravox_v06_qwen3_32b} backbones in Table~\ref{tab:bfclv3}. We organize Ultravox models in a different section since they are turn-based models, and we label the best score in bold for each group. Our system with the 7B backend substantially outperforms Ultravox-v0.6 Llama-3.1-8B in average score (71.7 vs.\ 43.3). For Ultravox-v0.6-Llama-3.1-8B, we use post-hoc type normalization to coerce string-typed JSON arguments to their schema types. The score of 0.0 on Irrelevance is because it invokes the single offered function on all 240 prompts.
For GPT-realtime, we use semantic VAD and feed the user speech turn to the model to generate the tool-call request. As shown in Table \ref{tab:bfclv3}, our 30B backend setup performs close to GPT-realtime on some subsets; however, the largest gap is on Parallel-Multiple, followed by Irrelevance.
We also compute tool-call token recall as the fraction of positive tool-call utterances for which the frontend correctly fires the delegation token. Our frontend achieves recall of 97.2\%, 92.0\%, 95.0\%, and 93.5\% on Simple, Multiple, Parallel, and Parallel Multiple, respectively.

\subsubsection{ Full-Duplex-Bench v3}

\begin{table}[!t]
\centering
{\footnotesize\caption{FDB3 results; ``--'' is N/A. Boldface marks group bests. Ours-235B-extASR denotes our system with the Qwen3-235B-A22B backend and external ASR. $^{*}$Judged with no timing.}\label{tab:fdb3}}
\footnotesize
\setlength{\tabcolsep}{0.75pt}
\setlength{\aboverulesep}{0.2ex}
\setlength{\belowrulesep}{0.2ex}
\renewcommand{\arraystretch}{1.0}
\begin{tabular}{@{}lrrrrrrr@{}}
\toprule
{\scriptsize Model} & {\scriptsize Tool-acc$\uparrow$} & {\scriptsize Arg-acc$\uparrow$} & {\scriptsize Pass@1$\uparrow$} & {\scriptsize Res-Q$\uparrow$} & {\scriptsize TT$\uparrow$} & {\scriptsize Inter$\downarrow$} & {\scriptsize Filler$\downarrow$} \\
\midrule
GPT-realtime-mini \cite{openai_gpt_realtime_mini} & 77.4 & 58.2 & 51.0 & 62.0 & 93.0 & 14.2 & 17.2 \\
GPT-realtime & \textbf{88.7} & \textbf{71.6} & \textbf{61.0} & \textbf{74.7} & \textbf{95.0} & \textbf{13.5} & \textbf{12.1} \\
\midrule
Gemini-3.5 Flash \cite{gemini_35_flash} & \textbf{97.0} & \textbf{72.5} & \textbf{65.0} & \textbf{88.0} & -- & -- & \textbf{1.0}$^{*}$ \\
UV0.6-8B & 59.1 & 39.0 & 11.0 & 19.0 & -- & -- & 2.0$^{*}$ \\
UV0.6-32B & 84.3 & 52.2 & 45.0 & 71.0 & -- & -- & 12.0$^{*}$ \\
\midrule
Ours-30B-extASR & \textbf{74.6} & 52.8 & 44.0 & 54.0 & 100.0 & 54.0 & 84.8 \\
Ours-235B-extASR & 71.7 & \textbf{55.2} & \textbf{48.0} & \textbf{67.0} & \textbf{100.0} & \textbf{51.0} & \textbf{83.3} \\
\bottomrule
\end{tabular}
\end{table}

Whereas BFCL contains continuous and fluent TTS-generated user speech,
Full-Duplex-Bench v3 (FDB3) \cite{fdbenchv3tool} consists
entirely of naturalistic real human recordings. 
The corpus contains $100$ scenarios from $12$ speakers,
recorded with everyday built-in microphones in environments
ranging from quiet rooms to mild background noise.
Mock APIs are used to generate tool-call results. 
In evaluation, we follow the official FDB3 setup to use GPT-4o as the LLM judge \cite{fdb3_hf}, along with the original system prompts and tool descriptions. In this evaluation, we stream user speech to the frontend chunk by chunk as in live conversation. Since our frontend is an STT model, we use agent text to compute Res-Q, TT, and Inter. We also reproduced the other models' results using their text outputs in Table \ref{tab:fdb3}. Thinking is set to medium for Gemini-3.5 Flash.

Table~\ref{tab:fdb3} reports tool-selection accuracy (Tool-acc), argument accuracy (Arg-acc), response quality (Res-Q), end-to-end task completion (Pass@1), turn-taking rate (TT), interruption rate (Inter), and filler rate (Filler). We remove end-to-end latency because our backends may run either as local vLLM instances or through cloud APIs, and their latencies are not comparable. The turn-taking latency of our frontend model is evaluated separately in Sect. \ref{sec:tt_intel}.

As shown in Table \ref{tab:fdb3}, our systems fulfill the user's request (Res-Q) at around 54--67\%. A larger backend (Qwen3-235B-A22B) performs better as expected. Our model achieves 100\% turn-taking.
Our high \emph{Filler} rate is by design: the frontend emits a short hold phrase (``Let me pull that up'') before the backend executes the tool call. Our high \emph{Inter} rate has a different cause: the frontend typically responds at pauses during the user's disfluency with a backchannel (e.g., ``Okay, I am here.'') before any tool call. Because it keeps listening while speaking, the eventual tool call usually still sees the complete user request (\emph{Res-Q} up to 67.0).

We compare our systems to open- and closed-source models in Table \ref{tab:fdb3}, where the Ultravox and Gemini models are grouped into a section representing turn-based speech models, and therefore the \emph{TT} and \emph{Inter} metrics are not applicable. Our models perform substantially better than Ultravox-v0.6 Llama-3.1-8B because the model produced no response in many scenarios. Ultravox-v0.6 Qwen-3-32B performs better than our systems on some metrics because it is a turn-based model and will not emit premature tool calls. For GPT-realtime and GPT-realtime-mini, we use semantic VAD and take their text output for evaluation. Compared with them, our results with the Qwen3-235B-A22B backend are similar to those of GPT-realtime-mini based on Pass@1 and Res-Q, while GPT-realtime performs best on almost all metrics among the duplex speech models evaluated.

\compactsubsection{Voice Agent Tasks}
\label{sec:eva}

Finally, we evaluate on EVA-Bench \cite{evabench}, an end-to-end framework for evaluating voice-agent tasks on grounded, multi-domain customer-service tasks. Our model's EVA results are generated using the following setup. We simulate the caller using EVA's \texttt{LiteLLMClient} and use GPT-5.2 to role-play the caller. To adapt to our infrastructure and improve inference availability, we synthesize user text turns with Chatterbox-TTS \cite{chatterboxflash}, and the synthesized user speech is streamed chunk by chunk to the frontend implemented using Triton Inference Server \cite{tritonserver}. We use GPT-5.2 as the LLM judge for all metrics. To focus on evaluating the quality of our frontend's STT text output and to compare with other text and STT models, we directly return agent text to the simulated user rather than transcripts of synthesized speech for both our models and the comparison models in Table \ref{tab:eva_backends}. The API's default reasoning is used for GPT-realtime-2 and thinking is also turned on for GPT-5.2 and Gemini 3.1 Flash Lite.

For metrics, we report all 213 EVA-Bench scenarios across the airline, ITSM, and medical HR domains. Since we use text outputs from the agent, we adapt \textbf{EVA-A} and \textbf{EVA-X} metrics for evaluation. EVA-A is the average of \emph{Task} and \emph{Faith}(fulness), and we remove \emph{Speech Fidelity}, which measures TTS quality, since our frontend outputs agent text. EVA-X is the average of \emph{Prog}(ress), \emph{Concise}(ness), and \emph{Speak}(ability). We add Speakability to measure how friendly the agent text is as input to the TTS model.

\begin{table}[!t]
\centering
{\footnotesize\caption{EVA-Bench \cite{evabench} results. Values are percentages except TT (s). Q3O-30B and G3.1-FL denote Qwen3-Omni-30B-A3B-Instruct and Gemini 3.1 Flash Lite, respectively.}\label{tab:eva_backends}}
\scriptsize
\renewcommand{\arraystretch}{1.0}
\setlength{\tabcolsep}{0.8pt}
\setlength{\aboverulesep}{0.2ex}
\setlength{\belowrulesep}{0.2ex}
\resizebox{\columnwidth}{!}{%
\begin{tabular}{@{}lrrrrrrrr@{}}
\toprule
Model & EVA-A$\uparrow$ & EVA-X$\uparrow$ & Task$\uparrow$ & Faith$\uparrow$ & Prog.$\uparrow$ & Concise$\uparrow$ & Speak$\uparrow$ & TT$\downarrow$ \\
\midrule
GPT-5.2 (text) & \textbf{63.3} & \textbf{80.8} & \textbf{78.4} & \textbf{48.1} & \textbf{73.2} & \textbf{77.6} & \textbf{91.6} & -- \\
\midrule
GPT-RT-mini & 33.1 & 71.2 & 37.1 & 29.1 & 46.5 & 77.5 & 89.7 & \textbf{1.35} \\
GPT-RT2 & \textbf{59.4} & \textbf{75.7} & \textbf{68.1} & \textbf{50.7} & \textbf{63.9} & 72.7 & 90.5 & 1.65 \\
\midrule
Q3O-30B & 29.1 & 72.7 & 32.5 & 25.7 & 46.9 & 77.5 & 93.7 & -- \\
G3.1-FL & \textbf{45.6} & \textbf{75.3} & \textbf{57.1} & \textbf{34.2} & 53.1 & \textbf{78.9} & \textbf{93.8} & -- \\
UV0.6-8B~\cite{ultravox_v06_llama31_8b} & 8.3 & 58.8 & 11.8 & 4.7 & 16.5 & 68.5 & 91.5 & -- \\
UV0.6-32B~\cite{ultravox_v06_qwen3_32b} & 31.2 & 54.5 & 36.2 & 26.3 & \textbf{55.2} & 68.2 & 40.1 & -- \\
\midrule
Ours-30B-extASR & 32.4 & 63.5 & 40.4 & 24.4 & 39.2 & 65.7 & 85.6 & \textbf{2.80} \\
Ours-235B-extASR & \textbf{46.6} & \textbf{74.7} & \textbf{57.3} & \textbf{35.9} & \textbf{53.5} & \textbf{77.6} & \textbf{93.0} & 3.14 \\
\bottomrule
\end{tabular}
}
\end{table}
\vspace{-0.5ex}

In Table \ref{tab:eva_backends}, we compare EVA performance across the text-only model (GPT-5.2), GPT-realtime-mini \cite{openai_gpt_realtime_mini}, GPT-realtime2 \cite{openai_gpt_realtime2}, turn-based speech models, and our frontend--backend systems. GPT-5.2 text-only is presented as a topline model, where we directly pass user text to the model to obtain the agent text response. This ideal setting performs best, with the highest EVA-A and EVA-X scores and a task-completion ratio of 78.4\%. We also run GPT-realtime-mini and GPT-realtime2 in streaming mode using the semantic VAD setup and return text outputs directly to the simulated user. GPT-realtime2 performs substantially better than the mini on EVA-A.
Our model with the Qwen3-30B-A3B backend performs similarly to GPT-realtime-mini on EVA-A, and the Qwen3-235B-A22B backend achieves substantially better EVA-A and EVA-X scores than GPT-realtime-mini but still lags behind GPT-realtime2 on EVA-A.

For turn-based speech models in Table \ref{tab:eva_backends}, we directly use the full user speech as input. Our system with the Qwen3-235B-A22B backend performs similarly to Gemini 3.1 Flash Lite and outperforms Ultravox-v0.6 Qwen-3-32B \cite{ultravox_v06_qwen3_32b}. Compared with Qwen3-Omni-30B-A3B-Instruct, our system with the Qwen3-30B-A3B backend achieves higher EVA-A and task completion. For Qwen3-235B-A22B, the conciseness and speakability scores are better than those of GPT-realtime2, but task completion and faithfulness lag behind. In a more detailed analysis, we find that the Qwen3-235B-A22B backend achieves higher task completion in the airline domain than GPT-realtime2 (72\% vs. 54\%) but underperforms in the ITSM (56.3\% vs. 75\%) and Medical HR (49.4\% vs. 69.9\%) domains. In the latter two scenarios, the model needs a chain of up to 6--8 successful tool calls to complete a task. This requires reliable tool-call detection from the frontend over a long conversation.

We also report the median \emph{turn-taking latency} (TT) in Table \ref{tab:eva_backends} for realtime models. For our systems, TT is the median time from the end of user speech to the onset of the non-filler agent response including both frontend-only and backend-involved tool-call turns. We achieved 2.80\,s and 3.14\,s for the 30B and 235B backends, respectively. The latencies are higher than GPT-realtime2 and GPT-realtime-mini partly because we wait for the LangGraph backend agent and LLM to compute the full result for prefilling. We did not optimize the prefill or serving infrastructure to specifically reduce latency in this work. However, we note that we train the frontend to acknowledge the user with a filler while they wait, and the filler response latencies are 0.8\,s and 0.72\,s for 30B and 235B backends, respectively, for tool-call turns.

\compactsubsection{Turn-Taking and Intelligence}
\label{sec:tt_intel}

\begin{table}[!t]
\centering
{\footnotesize\caption{Turn-taking and barge-in (BI) on our internal benchmark.}\label{tab:tt}}
\setlength{\tabcolsep}{2pt}
\setlength{\aboverulesep}{1.2pt}
\setlength{\belowrulesep}{1.2pt}
\renewcommand{\arraystretch}{1.13}
\footnotesize
\begin{tabular*}{\columnwidth}{@{\extracolsep{\fill}}lccc@{}}
\toprule
Model & Pr$\uparrow$/Rec$\uparrow$ (\%) & Lat.$\downarrow$ (ms) & BI Acc.$\uparrow$/Lat.$\downarrow$ (\%/ms) \\
\midrule
Baseline (no TC) & \textbf{85}/\textbf{94} & \textbf{423} & \textbf{100}/403 \\
Ours (frontend) & 82/91 & 438 & 99/\textbf{395} \\
\bottomrule
\end{tabular*}
\end{table}

\begin{table}[t]
\centering
{\footnotesize\caption{Spoken-language intelligence. OpenBookQA reports accuracy (\%); AE and CE use a 5-point scale.}\label{tab:intel}}
\setlength{\tabcolsep}{6pt}
\setlength{\aboverulesep}{1.2pt}
\setlength{\belowrulesep}{1.2pt}
\renewcommand{\arraystretch}{1.13}
\footnotesize
\begin{tabular}{lccc}
\toprule
Model & OpenBookQA (\%)$\uparrow$ & AE (/5)$\uparrow$ & CE (/5)$\uparrow$ \\
\midrule
Baseline (no TC) & \textbf{65.0} & 3.54 & \textbf{2.87} \\
Ours (frontend) & 64.7 & \textbf{3.61} & 2.36 \\
\bottomrule
\end{tabular}
\end{table}

\begin{table}[t]
\centering
{\footnotesize\caption{ASR word error rate for the Open ASR Leaderboard.}\label{tab:asr}}
\setlength{\tabcolsep}{2.5pt}
\setlength{\aboverulesep}{1.2pt}
\setlength{\belowrulesep}{1.2pt}
\renewcommand{\arraystretch}{1.17}
\scriptsize
\resizebox{\columnwidth}{!}{%
\begin{tabular}{lccccccccc}
\toprule
Model & LS-C & LS-O & TED & Vox & Earn & AMI & Giga & SPGI & Avg \\
\midrule
Baseline (no TC) & \textbf{3.93} & 8.59 & \textbf{5.89} & \textbf{8.90} & \textbf{20.00} & \textbf{20.45} & \textbf{13.45} & 5.15 & \textbf{10.80} \\
Ours (frontend) & 4.16 & \textbf{8.18} & 6.71 & 9.90 & 21.52 & 21.66 & 14.64 & \textbf{4.98} & 11.47 \\
\bottomrule
\end{tabular}%
}
\end{table}

\begin{table}[!t]
\centering
{\footnotesize\caption{Full-Duplex-Bench v1 \cite{fdbench} results. Pause reports Candor TOR. GPT score is out of 5.}\label{tab:fdbv1}}
\setlength{\tabcolsep}{2pt}
\setlength{\aboverulesep}{1.2pt}
\setlength{\belowrulesep}{1.2pt}
\setlength{\cmidrulesep}{0.5pt}
\renewcommand{\arraystretch}{1.17}
\scriptsize
\resizebox{\columnwidth}{!}{%
\begin{tabular}{lcccccc}
\toprule
& \multicolumn{2}{c}{Smooth TT (Candor)} & Pause & \multicolumn{3}{c}{User Interruption} \\
\cmidrule(lr){2-3} \cmidrule(lr){4-4} \cmidrule(lr){5-7}
Model & TOR (\%)$\uparrow$ & Lat. (ms)$\downarrow$ & TOR (\%)$\downarrow$ & TOR (\%)$\uparrow$ & GPT score$\uparrow$ & Lat. (ms)$\downarrow$ \\
\midrule
Baseline (no TC) & 93 & 221 & \textbf{53.2} & \textbf{95.5} & 3.94 & 590 \\
Ours (frontend) & \textbf{100} & \textbf{92} & 68.2 & 89.5 & \textbf{4.04} & \textbf{349} \\
\bottomrule
\end{tabular}
}
\end{table}

Lastly, we assess whether tool-call training affects regular duplex conversation quality by comparing our frontend with a baseline trained from the same initialization and under the same settings, but without tool-call data. We evaluate the frontend on an internal
multi-turn conversation set \cite{hu2026enable}, VoiceBench tasks \cite{voicebench}, the Open ASR Leaderboard \cite{openasr_leaderboard}, and FDB-v1 \cite{fdbench}. 
In Tables~\ref{tab:tt}, \ref{tab:intel}, \ref{tab:asr}, and \ref{tab:fdbv1}, we compare the proposed model's performance on turn-taking, intelligence, streaming ASR, and FDB-v1 against a baseline model without tool-call (TC) training. 
We follow \cite{hu2026enable} to compute the metrics for the internal set. 
Overall, our frontend model performs slightly worse in turn-taking precision and recall while performing similarly on the other metrics in Table~\ref{tab:tt}. It performs similarly on OpenBookQA and AlpacaEval (AE) but scores lower on CommonEval (CE), as shown in Table~\ref{tab:intel}. The streaming ASR WER increases from 10.80\% to 11.47\% (Table~\ref{tab:asr}), presumably because the added tool-call training data are mostly synthetic. On FDB-v1 in Table~\ref{tab:fdbv1}, the model generally becomes more responsive, with a higher TOR rate for smooth TT and lower latency, but it also performs worse on pause handling. This can be addressed by adding more training data with natural pauses, which are not present in the current SFT training.

\compactsection{Conclusion}

We presented a modular frontend-backend architecture that adds tool use to full-duplex speech models through delegation and prefill-and-repeat. Across speech BFCL, FDB3, and EVA-Bench, the approach achieves high delegation recall and competitive task completion while largely preserving turn-taking and ASR performance. Robustness to natural pauses and long tool-call sequences remains an important direction for future work.

\compactsection*{Acknowledgment}
We thank Lily Lee, Nourchene Ferchichi, Harishchandra Dubey, Yuanhang Su, Aditya Malte, Zijia Chen, Travis Bartley, Praise Manzi, Hayley Ross, and Yoshi Suhara for their efforts, contributions, and support throughout this project.

Claude Opus 4.8 and Codex with GPT-5.5 were used only to format tables and references and to correct grammatical errors throughout the paper.

\clearpage
\bibliographystyle{IEEEbib}
\begingroup
\apptocmd{\thebibliography}{%
  \renewcommand{\baselinestretch}{1}%
  \fontsize{9}{9.1}\selectfont
  \setlength{\itemsep}{0pt}%
  \setlength{\parsep}{0pt}%
  \setlength{\parskip}{0pt}%
}{}{}
\bibliography{refs}
\endgroup

\end{document}